\documentclass[11pt,a4paper]{article}
\usepackage[hyperref]{acl2020}
\usepackage{times}
\usepackage{latexsym}
\usepackage{amsmath}
\usepackage{fdsymbol}
\usepackage{enumerate}
\usepackage{multirow}
\usepackage{enumitem}
\usepackage{lipsum}
\usepackage{url}
\usepackage{graphicx}
\usepackage{lineno}
\usepackage{natbib}
\usepackage{xcolor}

\usepackage{caption}
\usepackage{microtype}

\aclfinalcopy 

\title{Orthogonal Relation Transforms with Graph Context Modeling for Knowledge Graph Embedding}
\author{Yun Tang,  Jing Huang, Guangtao Wang, Xiaodong He, Bowen Zhou \\
  JD AI Research \\
  \texttt{\small\{yun.tang,jing.huang,guangtao.wang,xiaodong.he,bowen.zhou\}@jd.com} \\
}

\date{}

\begin{document}
\maketitle
\begin{abstract}
Distance-based knowledge graph embeddings have shown substantial improvement on the knowledge graph link prediction task, from {\em TransE} to the latest state-of-the-art {\em RotatE}. However, complex relations such as N-to-1, 1-to-N and N-to-N still remain challenging to predict. 
In this work, we propose a novel distance-based approach 
for knowledge graph link prediction. 
First we extend the {\em RotatE} from 2D complex domain to high dimensional space with orthogonal transforms to model relations. The orthogonal transform embedding for relations keeps the capability for modeling symmetric/anti-symmetric, inverse and compositional relations while achieves better modeling capacity. 
Second, the graph context 
is integrated into distance scoring functions directly. 
Specifically, graph context is explicitly modeled via two directed context representations. Each node embedding in knowledge graph is augmented with two context representations, which are computed from the neighboring outgoing and incoming nodes/edges respectively. 
The proposed approach improves prediction accuracy on the difficult N-to-1, 1-to-N and N-to-N cases. Our experimental results show that it achieves state-of-the-art results 
on two common benchmarks FB15k-237 and WNRR-18,
especially on FB15k-237 which has many high in-degree nodes.
\end{abstract}

\section{Introduction}\label{sec:intr}

\begin{figure*}[!ht]
    \centering
    \includegraphics[width=.9\textwidth]{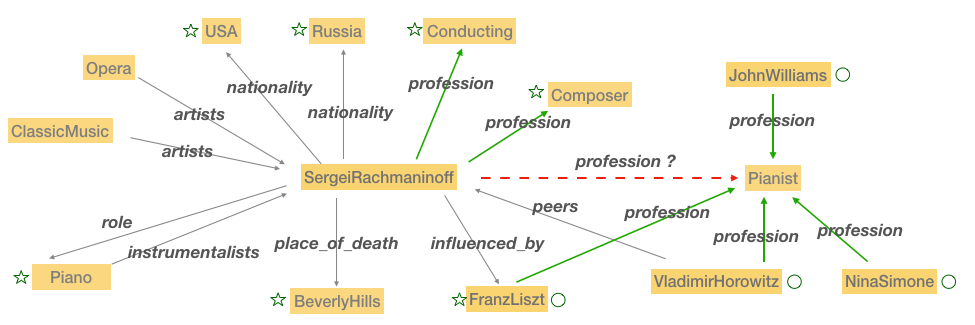}
    \caption{Snapshot of knowledge graph in FB15k-237. Entities are represented as golden blocks. }
    \label{fig:kg}
\end{figure*} 

Knowledge graph is a multi-relational graph whose nodes represent entities and edges denote relationships between entities. 
Knowledge graphs store facts about people, places and world from various sources. Those facts are kept as triples (head entity, relation, tail entity) and denoted as $(h, r, t)$.
A large number of knowledge graphs, such as Freebase~\cite{bollacker2008freebase}, DBpedia~\cite{auer2007dbpedia}, NELL~\cite{carlson2010toward} and YAGO3~\cite{mahdisoltani2013yago3}, have been built over the years and successfully applied to many domains such as recommendation and question answering~\cite{Bordes2014QuestionAW,Zhang2016CollaborativeKB}.
However, these knowledge graphs need to be updated with new facts periodically.
Therefore many knowledge graph embedding methods have been proposed for link prediction that is used for knowledge graph completion.

Knowledge graph embedding represents entities and relations in continuous vector spaces.
Started from a simple and effective approach called {\it TransE}~\cite{bordes2013translating}, many knowledge graph embedding methods have been proposed, such as {\it TransH}~\cite{wang2014knowledge}, {\it DistMult}~\cite{yang2014distmult}, {\it ConvE}~\cite{dettmers2017conve} to the latest {\it RotatE}~\cite{Sun2019RotatEKG} and {\it QuatE }~\cite{zhang2019quaternion}.

Though much progress has been made, 1-to-N, N-to-1, and N-to-N relation predictions \cite{bordes2013translating,wang2014knowledge} still remain challenging. In Figure~\ref{fig:kg}, relation ``profession'' demonstrates an N-to-N example and the corresponding edges are highlighted as green. Assuming the triple (SergeiRachmaninoff, Profession, Pianist) is unknown.
The link prediction model takes ``SergeiRachmaninoff'' and relation ``Profession'' and rank all entities in the knowledge graph to predict ``Pianist''. Entity ``SergeiRachmaninoff'' connected to multiple  entities as head entity via relation ``profession'', while ``Pianist'' as a tail entity also reaches to multiple entities through relation ``profession''. It makes the N-to-N prediction hard because the mapping from certain entity-relation pair could lead to multiple different entities.
Same issue happens with the case of 1-to-N and N-to-1 predictions.

The recently proposed {\it RotatE} \cite{Sun2019RotatEKG} models each relation as a 2-D rotation from the source entity to the target entity.
The desired properties for relations include symmetry/antisymmery, inversion and composition which have been demonstrated to be useful for link prediction in knowledge graph. Many existing methods model one or a few of these relation patterns, while {\it RotatE} naturally handles all these relation patterns. 
In addition, the entity and relation embeddings are divided into multiple groups (for example, $1000$ 2-D rotations are used in \cite{Sun2019RotatEKG}). 
Each group is modeled and scored independently. The final score is computed as the summation of all these scores, which can be viewed as an ensemble of different models and further boost the performance of link prediction. However, {\it RotatE} is limited to 2-D rotations and thus has limited modeling capacity. In addition, {\it RotatE} does not consider graph context, which is helpful in handling 1-to-N, N-to-1, and N-to-N relation prediction.
 



In this work, a novel distance-based knowledge graph embedding called orthogonal transform embedding ({\it OTE}) with graph context is proposed to alleviate the 1-to-N, N-to-1 and N-to-N issues, while keeps the desired relation patterns as {\it RotatE}.
First, we employ orthogonal transforms to represent relations in high dimensional space for better modeling capability. The Orthogonal transform embedding also models the symmetry/antisymmery, inversion and compositional relation patterns just as {\it RotatE} does. {\it RotatE} can be viewed as an orthogonal transform in 2D complex space.

Second, we integrate graph context directly into the distance scoring, which is helpful to predict 1-to-N, N-to-1 and N-to-N relations. For example,
from the incomplete knowledge graph, people find useful context information, such as (SergeiRachmaninoff, role, Piano) and (SergeiRachmaninoff, Profession, Composer) in Figure~\ref{fig:kg}.
In this work, each node embedding in knowledge graph is augmented with two graph context representations, computed from the neighboring outgoing and incoming nodes respectively.
Each context representation is 
computed based on the embeddings of the neighbouring nodes and the corresponding relations connecting to these neighbouring nodes.
These context representations are  used as part of the distance scoring function to measure the plausibility of the triples during training and inference. 
We show that {\em OTE } together with graph context modeling performs consistently better than {\em RotatE} on the standard benchmark FB15k-237 and WN18RR datasets.



 In summary, our main contributions include:
 \begin{itemize}[leftmargin=10pt]
     \item A new orthogonal transform embedding {\it OTE}, is proposed to extend {\it RotatE} from 2D space to high dimensional space, which also models symmetry/antisymmery, inversion and compositional relation patterns;
     \item A directed graph context modeling method is proposed to integrate knowledge graph context (including both neighboring entity nodes and relation edges) into the distance scoring function;
     \item Experimental results of {\it OTE} on standard benchmark FB15k-237 and WN18RR datasets show consistent improvements over {\it RotatE}, the state of art distance-based embedding model, especially on FB15k-237 with many high in-degree nodes. On WN18RR our results achieve the new state-of-the-art performance.
 \end{itemize}

\section{Related work}\label{sec:related}
\subsection{Knowledge Graph Embedding}
Knowledge graph embedding could be roughly categorized into two classes~\cite{Wang2017KnowledgeGE}: distance-based models and semantic matching models. Distance-based model is also known as additive models, since it projects head and tail entities into the same embedding space and the distance scoring between two entity embeddings is used to measure the plausibility of the given triple. {\it TransE}~\cite{bordes2013translating} is the first and most representative translational distance model. A series of work is conducted along this line such as {\it TransH}~\cite{wang2014knowledge}, {\it TransR}~\cite{lin2015learning} and {\it TransD}~\cite{Ji2015KnowledgeGE} etc. {\it RotatE}~\cite{Sun2019RotatEKG} further extends the computation into complex domain and is currently the state-of-art in this category. On the other hand, Semantic  matching  models usually take multiplicative score functions to compute the plausibility of the given triple, such as {\it DistMult}~\cite{yang2014distmult}, {\it ComplEx}~\cite{trouillon2016complex}, {\it ConvE}~\cite{dettmers2017conve}, {\it TuckER}~\cite{balazevic-etal-2019-tucker} and {\it QuatE}~\cite{zhang2019quaternion}. {\it ConvKB}~\cite{nguyen2017novel} and {\it CapsE}~\cite{Nguyen2019CapsE} further took the triple as a whole, and fed head, relation and tail embeddings into convolutional models or capsule networks.

The above knowledge graph embedding methods focused on modeling individual triples.
However, they ignored knowledge graph structure and did not take advantage of context from neighbouring nodes and edges. This issue inspired the usage of graph neural networks~\cite{kipf2016semi,velivckovic2017graph} for graph context modeling. 
Encoder-decoder framework was adopted in~\cite{Schlichtkrull2017ModelingRD,Shang2019EndtoEndSC,Bansal2019A2NAT}. The knowledge graph structure is first encoded via graph neural networks and the output with rich structure information is passed to the following graph embedding model for prediction. The graph model and the scoring model could be end-to-end trained together, or the graph encoder output was only used to initialize the entity embedding~\cite{Nathani2019LearningAE}. We take another approach in this paper: we integrate the graph context directly into the distance scoring function.

\subsection{Orthogonal Transform}
Orthogonal transform is considered to be more stable and efficient for neural networks~\cite{Saxe2013ExactST,Vorontsov2017OnOA}. 
However, to optimize a linear transform with orthogonal property reserved is not straightforward.  Soft constraints could be enforced during optimization to encourage the learnt linear transform close to be orthogonal. \citet{Bansal2018CanWG} extensively compared different orthogonal regularizations and find regularizations make the training faster and more stable in different tasks. On the other hand, some work has been done to achieve strict orthogonal during optimization by applying special gradient update scheme. \citet{Harandi2016GeneralizedB} proposed a Stiefel layer to guarantee fully connected layers to be orthogonal by using Reimannian gradients. \citet{Huang2017OrthogonalWN} consider the estimation of orthogonal matrix as an optimization over multiple dependent stiefel manifolds problem and  solve it via eigenvalue decomposition on a proxy parameter matrix.
\citet{Vorontsov2017OnOA} applied hard constraint on orthogonal transform update via Cayley transform. In this work, we construct the orthogonal matrix via Gram Schmidt process and the gradient is calculated automatically through autograd mechanism in PyTorch~\cite{paszke2017automatic}.




\section{Our Proposed Method}

We consider knowledge graph as a collection of triples $\mathcal{D}=\{(h,r,t)\}$ with $V$ as the graph node set, and $R$ as the graph edge set. Each triple has a head entity $h$ and tail entity $t$, where $h,t\in V$. Relation $r\in R$ connects two entities with direction from head to tail.
As discussed in the introduction section, 1-to-N, N-to-1 and N-to-N relation prediction \cite{bordes2013translating,wang2014knowledge} are difficult to deal with. They are addressed in our proposed approach by: 1) orthogonal relation transforms that operate on groups of embedding space. Each group is modeled and scored independently, and the final score is the sum of all group scores. Hence, each group could address different aspects of entity-relation pair and alleviate the 1-to-N and N-to-N relation mapping issues; and 2) directed graph context to integrate knowledge graph structure information to reduce the ambiguity. 

Next, we first briefly review {\it RotatE} that motivates our orthogonal transform embedding ({\em OTE}), and then describe the proposed method in details.

\subsection{RotatE}
{\em OTE} is inspired by {\it RotatE}~\cite{Sun2019RotatEKG}.
In {\it RotatE}, the distance scoring is done via Hadamard production (element-wise) defined on the complex domain. Given a triple $(h,r,t)$, the corresponding embedding are $e_h$, $\theta_r$, $e_t$, where $e_h$ and $e_t \in \mathcal{R}^{2d}$, $\theta_r \in \mathcal{R}^{d}$,  and $d$ is the embedding dimension.  For each dimension $i$, $e[2i]$ and $e[2i+1]$ are corresponding real and imaginary components. The projection $\tilde{e}_t$ of $t$ from corresponding relation and head entities is conducted as an orthogonal transform as below:
\begin{eqnarray}
  \begin{bmatrix}\tilde{e}_t[2i]\\ \tilde{e}_t[2i\!+\!1]\end{bmatrix} \!\!\!\!&=&\!\!\! M_r(i)\begin{bmatrix}e_h[2i]\\ e_h[2i\!+\!1]\end{bmatrix} \nonumber\\ 
  \!\!\!\!&=&\!\!\! \begin{bmatrix}
    \cos{\theta_r(i)} &\! -\sin{\theta_r(i)} \\
    \sin{\theta_r(i)} &\! \cos{\theta_r(i)}
    \end{bmatrix}\begin{bmatrix}e_h[2i]\\ e_h[2i\!+\!1]\end{bmatrix} \nonumber
\end{eqnarray}
where $M_r(i)$ is a 2D orthogonal matrix derived from $\theta_r$ .

Though {\it RotatE} is simple and effective for knowledge graph link prediction, it is defined in $2D$ complex domain and thus has limited modeling capability. 
A natural extension is to apply similar operation on a higher dimensional space.

\subsection{Orthogonal Transform Embedding (OTE)} \label{sec:ote}
We use $e_h$, $M_r$, $e_t$ to represent embeddings of head, relation and tail entity, where $e_h$, $e_t \in \mathcal{R}^d$,  and $d$ is the dimension of the entity embedding. The entity embedding $e_{x}$, where $x= \{h, t\}$, is further divided into $K$ sub-embeddings, e.g., $e_{x} = [e_{x}(1);\cdots;e_{x}(K)]$, where $e_{x}(i)\in\mathcal{R}^{d_s}$ and $d=K\cdot d_s$. $M_r$ is a collection of $K$ linear transform matrix $M_r = \{ M_r(1),\cdots,M_r(K)\}$, and $M_r(i)\in\mathcal{R}^{d_s\times d_s}$. 

For each sub-embedding $e_{t}(i)$ of tail $t$, we define the projection from $h$ and $r$ to $t$ as below:
\begin{equation}
    \tilde{e}_t(i) = f_{i}(h, r) = \phi(M_r(i)) e_h(i) \label{equ:hr_t_g}
\end{equation}
where $\phi$ is the Gram Schmidt process (see details in Section~\ref{sec:GS}) applied to square matrix $M_r(i)$. The output transform $\phi(M_r(i))$ is an orthogonal matrix derived from $M_r(i)$. 
$\tilde{e}_t$ is the concatenation of all sub-vector $\tilde{e}_t(i)$ from Eq.~\ref{equ:hr_t_g}, e.g., $\tilde{e}_t =f(h, r) = [\tilde{e}_t(1);\cdots;\tilde{e}_t(K)]$. 
The $L_2$ norm of $e_h(i)$ is preserved after the orthogonal transform.  
We further use a scalar tensor $s_r(i) \in \mathcal{R}^{d_s}$ to scale the $L_2$ norm of each group of embedding separately. Eq.~\ref{equ:hr_t_g} is re-written as
\begin{equation}
    \tilde{e}_t(i) = diag(\exp(s_r(i))) \phi(M_r(i))  e_h(i) \label{equ:hr_t_g_s}
\end{equation}

Then, the corresponding distance scoring function is defined as
\begin{equation}
    d((h,r),t) = \sum_{i=1}^K ( ||\tilde{e}_t(i) - e_t(i)|| )  \label{equ:score_d_t}
\end{equation}




For each sub-embedding $e_{h}(i)$ of head $h$, we define the projection from $r$ and $t$ to $h$ as below:
\begin{eqnarray}
    \tilde{e}_h(i)=diag(\exp(-s_r(i)))\phi(M_r(i))^T e_t(i) \label{equ:hr_t_g_r}
\end{eqnarray}
where the reverse project from tail to head is simply transposing the $\phi(M_r(i))$ and reversing the sign of $s_r$. Then, the corresponding distance scoring function is defined as
\begin{equation}
    d(h,(r, t)) = \sum_{i=1}^K ( ||\tilde{e}_h(i) - e_h(i)||).  \label{equ:score_d_h}
\end{equation}




\subsection{Gram Schmidt Process}
\label{sec:GS}

We employ Gram-Schmidt process to orthogonalize a linear transform into an orthogonal transform (i.e., $\phi(M_r(i))$ in Section \ref{sec:ote}). The Gram-Schmidt process takes a set of tensor $S=\{v_1,\cdots,v_k\}$ for $k\leq {d_s}$ and generates an orthogonal set
$S'=\{u_1,\cdots,u_k\}$ that spans the same $k-$dimensional subspace of $\mathcal{R}^{d_s}$ as $S$. 
\begin{eqnarray}
    t_i &=& v_k - \sum_{j=1}^{k-1}\frac{\langle v_k,t_j\rangle}{\langle t_j,t_j\rangle}t_j \\
    u_i &=& \frac{t_i}{||t_i||} \label{gc_unit}
\end{eqnarray}
where $t_1=v_1$, $||t||$ is the $L_2$ norm of vector $t$ and $\langle v,t\rangle$ denotes the inner product of $v$ and $t$. 

Orthogonal transform has many desired properties,
for example, the inverse matrix is obtained by simply transposing itself. It also preserves the $L_2$ norm of a vector after the transform. 
For our work, we are just interested in its property to obtain inverse matrix by simple transposing. This saves the number of model parameters (see Table \ref{tab:ablation}).

It can be easily proved that {\em OTE} has the ability to model and infer all three types of relation patterns: symmetry/antisymmetry, inversion, and composition as {\em RotatE} does. The proof is listed in Appendix \ref{sec:appendix}.

It should be noted that, $M_r(i)$ is calculated every time in the neural networks forward computation to get orthogonal matrix $\phi(M_r(i))$, while the corresponding gradient is calculated and propagated back to $M_r(i)$ via autograd computation within PyTorch during the backward computation. 
It eliminates the need of special gradient update schemes employed in previous hard constraint based orthogonal transform estimations~\cite{Harandi2016GeneralizedB,Vorontsov2017OnOA}.
In our experiments, we initialize $M_r(i)$ to make sure they are with full rank\footnote{A real random matrix has full rank with probability 1 \cite{slinko2000generalization}. We use different random seeds to make sure the generated matrix is full rank.}. During training, we also keep checking the determinant of $M_r(i)$. We find the update is fairly stable that we don't observe any issues with sub-embedding dimensions varied from $5$ to $100$.





\subsection{Directed Graph Context}

The knowledge graph is a directed graph: valid triple $(h, r, t)$ does not mean $(t, r, h)$ is also valid. Therefore, for a given entity in knowledge graph, there are two kinds of context information: nodes that come into it and nodes that go out of it.
Specially, in our paper, for each entity $e$, we consider the following two context settings:
\begin{enumerate} [leftmargin=10pt]
\setlength{\itemsep}{0pt}
\setlength{\parskip}{0pt}
    \item If $e$ is a tail, all the (head, relation) pairs in the training triples whose tail is $e$ are defined as \textit{Head Relation Pair Context}.
    \item If $e$ is a head, all the (relation, tail) pairs in the training triples whose head is $e$ are defined as \textit{Relation Tail Pair Context}.
\end{enumerate}
Figure~\ref{fig:kg} demonstrates the computation of graph context for a testing triple (SergeiRachmaninoff, profession, Pianist). Edges for relation ``profession'' are colored as green. Entities marked with $\circ$ are head entities to entity ``Pianist'', and these entities and corresponding relations to connect ``Pianist'' form the head relation pair context of ``Pianist''.   While entities with $\smallwhitestar$ are tail entities for entity ``SergeiRachmaninoff''. Those entities and corresponding relations are the relation tail graph context of entity ``SergeiRachmaninoff''. 

\subsubsection{Head Relation Pair Context}
For a given tail $t$, all head-relation pairs $(h',r')$ of the triples with tail as $t$ are considered as its graph context and denoted as $Ng(t)$.

First, we compute the head-relation context representation $\tilde{e}_t^c$ as the average from all these pairs in $Ng(t)$ as below:

\begin{equation}
    \tilde{e}_t^c  = \frac{\sum_{(h',r')\in Ng(t)} f(h',r') + e_t}{|Ng(t)| + 1}\label{equ:cxt}
\end{equation}
where $e_t$ is the embedding of the tail $t$, $f(h', r')$ is the representation of $(h', r')$ induced from Eq.~\ref{equ:hr_t_g_s}. We use $e_t$ in Eq.~\ref{equ:cxt} to make the computation of context representation possible when $Ng(t)$ is empty. 
This can be viewed as a kind of additive smoothing for context representation computation.

Then, we compute the distance of the head-relation context of $t$ and the corresponding orthogonal transform based representation of a triple $(h,r,t)$ as follow.
 \begin{eqnarray}
    d_c((h,r),t) \!\!\!\!&=&\!\! \sum_{i=1}^K ( || \tilde{e}_t(i) - \tilde{e}_t^c(i)||) \label{equ:score_cxt_t}
\end{eqnarray}
where $\tilde{e}_t(i)$ is computed from Eq.~\ref{equ:hr_t_g_s}.

There is no new parameter introduced for the graph context modeling, since the message passing is done via {\it OTE} entity-relation project $f(h', r')$. The graph context can be easily applied to other translational embedding algorithms, such as {\it RotatE} and {\it TransE} etc, by replacing {\it OTE}.

\subsubsection{Relation Tail Pair Context}

For a given head $h$, all relation-tail pairs $(r', t')$ of the triples with head as $h$ are considered as its graph context and denoted as $Ng(h)$.

First, we compute the relation-tail context representation $\tilde{e}_h^c$ as the average from all these pairs in $Ng(h)$ as below:
\begin{equation}
    \tilde{e}_h^c  = \frac{\sum_{(r',t')\in Ng(h)} f(r', t') + e_h}{|Ng(h)| + 1}\label{equ:cxh}
\end{equation}
where $f(r', t')$ is computed from Eq.~\ref{equ:hr_t_g_r}.

Then, we compute the distance of the relation-tail context of $h$ and the corresponding orthogonal transform based representation of a triple $(h,r,t)$ as follow.
 \begin{eqnarray}
    d_c(h,(r,t)) \!\!\!\!&=&\!\! \sum_{i=1}^K ( || \tilde{e}_h(i) - \tilde{e}_h^c(i)||) \label{equ:score_cxt_h} 
\end{eqnarray}
where $\tilde{e}_h(i)$ is computed from Eq.~\ref{equ:hr_t_g_r}.

\subsection{Scoring Function}

We further combine all four distance scores (Eq.~\ref{equ:score_d_t}, Eq.~\ref{equ:score_d_h}, Eq.~\ref{equ:score_cxt_t} and Eq.~\ref{equ:score_cxt_h}) discussed above as the final distance score of the graph contextual orthogonal transform embedding ({\it  GC-OTE}) for training and inference
\begin{eqnarray}
    d_{all}(h,r,t) = d((h,r),t) + d_c(h,(r,t)) \nonumber\\
    + d(h,(r,t)) + d_c((h,r),t). \label{equ:score_d_all}
\end{eqnarray}

Therefore the full GC-OTE model can be seen as an ensemble of $K$ local GC-OTE models. This view provides an intuitive explanation for the success of GC-OTE.

\noindent\textbf{Optimization} Self-adversarial negative sampling loss \cite{Sun2019RotatEKG} is used to optimize the embedding in this work, 
\begin{eqnarray}
    L &=&  -\sum p(h',r,t')\log\sigma(d_{all}(h',r,t') - \gamma) \nonumber\\
     && -\log\sigma(\gamma - d_{all}(h,r,t))
\end{eqnarray}
where $\gamma$ is a fixed margin, $\sigma$ is sigmoid function, $(h',r,t')$ is negative 
triple, and $p(h',r,t')$ is the negative sampling weight defined in \cite{Sun2019RotatEKG}.

\section{Experiments}\label{sec:exp}

\subsection{Datasets}
Two commonly used benchmark datasets (FB15k-237 and WN18RR) are employed in this study to evaluate the performance of link prediction.

\noindent\textbf{FB15k-237} \cite{toutanova2015observed} dataset contains knowledge base relation triples and textual mentions of Freebase entity pairs. The knowledge base triples are a subset of the FB15K \cite{bordes2013translating}, originally derived from Freebase. The inverse relations are removed in FB15k-237.

\noindent\textbf{WN18RR} \cite{dettmers2017conve} is derived from WN18 \cite{bordes2013translating}, which is a subset of WordNet. WN18 consists of 18 relations and 40,943 entities. However, many text triples obtained by inverting triples from the training set. Thus WN18RR \cite{dettmers2017conve} is created to ensure that the evaluation dataset does not have test leakage due to redundant inverse relation. 

\begin{table}[!h]
	\begin{center}
		\setlength{\tabcolsep}{2pt}
		\begin{tabular}{|l|c|c|}
			\hline
			Dataset & FB15k-237 & WN18RR \\
			\hline
			Entities	 & 14,541	& 40,943   \\
			Relations	 & 237 & 11  \\
			Train Edges	 & 272,115 & 86,835  \\
			Val. Edges	 & 17,535 & 3,034  \\
			Test Edges	 & 20,466 & 3,134  \\
			\hline
		\end{tabular}
		\vspace{-5pt}
		\caption{Statistics of datasets.}\label{tbl:sample}
	\end{center}
\end{table}

Each dataset is split into three sets for: training, validation and testing, which is same with the setting of \cite{Sun2019RotatEKG}. The statistics of two data sets are summarized at Table \ref{tbl:sample}. Only triples in the training set are used to compute graph context.

\subsection{Evaluation Protocol}
Following the evaluation protocol in \cite{dettmers2017conve,Sun2019RotatEKG}, each test triple $(h,r,t)$ is measured under two scenarios: head focused $(?,r,t)$ and tail focused $(h,r,?)$. For each case, the test triple is ranked among all triples with masked entity replaced by entities in knowledge graph. Those true triples observed in either train/validation/test set except the test triple will be excluded during evaluation.  Top 1, 3, 10 (Hits@1, Hits@3 and Hits@10), and the Mean Reciprocal Rank (MRR) are reported in the experiments.

\subsection{Experimental Setup}

\textbf{Hyper-parameter settings} The hyper-parameters of our model are tuned by grid search during training process, including learning rate, embedding dimension $d$ and sub-embedding dimension $d_s$. In our setting, the embedding dimension is defined as the number of parameters in each entity embedding. Each entity embedding consists of $K$ sub-embeddings with dimension $d_s$, i.e., $d = K \times d_s$. There are two steps in our model training: 1) the model is trained with {\it OTE} or {\it RotatE} models, and 2) graph context based models are fine tuned on these pre-trained models. The parameter settings are selected by the highest MRR with early stopping on the validation set. We use the adaptive moment (Adam) algorithm \cite{kingma2014adam} to train the models.

Specially, for FB15k-237, we set embedding dimension $d=400$, sub-embedding dimension $d_s=20$, and the learning rates to $2e$-$3$ and $2e$-$4$ for pre-training and fine-tuning stages respectively; for WN18RR dataset, we set $d=400$, $d_s= 4$, and the learning rates to $1e$-$4$ and $3e$-$5$ for pre-training and fine-tuning stages. 

\noindent\textbf{Implementation} Our models are implemented by PyTorch and run on NVIDIA Tesla P40 Graphics Processing Units. 
The pre-training {\it OTE} takes 5 hours with 240,000 steps and fine-tuning {\it GC-OTE} takes 23 hours with 60,000 steps. Though, it takes more computation for graph context based model training, the inference could be efficient if both head and tail context representations are pre-computed and saved for each entity in the knowledge graph.  

\begin{table*}[!h]
    \small
    \tabcolsep=0.32cm
    \centering
    \begin{tabular}{c|c|c|c|c|c|c|c|c}
        \hline
        \multirow{2}{*}{Model} & \multicolumn{4}{c|}{\textbf{FB15k-237}} & \multicolumn{4}{c}{\textbf{WN18RR}}\\
        \cline{2-9}
         & MRR & H1 & H3 & H10 & MRR  & H1 & H3 & H10   \\
        \hline
        \hline
        TransE  & .294 & - & - & .465  & .226 & - & - & .501  \\
        \hline
        RotatE  & .338 & .241 & .375 & .533 & .476 & .428 & .492 & .571  \\
        \hline
        \hline
        DistMult & .241 & .155 & .263 & .419 & .43 & .39 & .44 & .49 \\ 
        \hline
        ComplEx & .247 & .158 & .275 & .428  & .44 & .41 & .46 & .51\\
        \hline
        ConvE  & .325 & .237 & .356 & .501  & .43 & .40 & .44 & .52 \\
        \hline
        QuatE & .348 & .248 & .382 & .550 & .488 & .438 & .508 & .582 \\
        \hline
        TurkER & .358 & .266 & .392 & .544 & .470 & .443 & .482 & .526 \\
        \hline
        \hline
        R-GCN+ & .249 & .151 & .264 & .417 & - & - & -& -  \\
        \hline
        SACN  & .352 & .261 & .385 & .536 & .47  & .43 & .48 & .54  \\
        \hline
        A2N & .317 & .232 & .348 & .486  & .45 & .42 & .46 & .51 \\
        \hline
        \hline
        {\bf OTE}  & { .351}  & { .258}  & { .388} & { .537}  & { .485} & { .437} & { .502}  & { .587}  \\\hline
        {\bf GC-OTE} & {\bf .361} & {\bf .267} & {\bf .396}  & {\bf .550} & {\bf .491} & {\bf .442} & {\bf .511}  & {\bf .583}  \\
        \hline
    \end{tabular}
    \caption{Link prediction for FB15k-237 and WN18RR on test sets.}\label{tab:main_results}
\end{table*}

\subsection{Experimental Results}
In this section, we first present the results of link prediction, followed by the ablation study and error analysis of our models.
\subsubsection{Results of Link Prediction}

Table~\ref{tab:main_results} compares the proposed models (\textit{OTE} and graph context based \textit{GC-OTE}) to several state-of-the-art models: including translational distance based {\textit{TransE}}~\cite{bordes2013translating}, {\textit{RotatE}}~\cite{Sun2019RotatEKG}; semantic matching based {\textit{DistMult}}~\cite{yang2014distmult}, {\textit{ComplEx}}~\cite{trouillon2016complex}, {\textit{ConvE}}~\cite{dettmers2017conve}, {\textit{TuckER}}~\cite{balazevic-etal-2019-tucker} and {\textit{QuatE}}~\cite{zhang2019quaternion}, and graph context information based {\textit{R-GCN+}}~\cite{Schlichtkrull2017ModelingRD}, {\textit{SACN}}~\cite{Shang2019EndtoEndSC} and {A2N}~\cite{Bansal2019A2NAT}. These baseline numbers are quoted directly from published papers.

From Table~\ref{tab:main_results}, we observe that: 1) on FB15k-237, {\it OTE} outperforms {\it RotatE}, and {\textit{GC-OTE}} outperforms all other models on all metrics.
Specifically MRR is improved from $0.338$ in {\textit{RotatE}}, to $0.361$, about $7$\% relative performance improvement. {\it OTE} which increases sub-embedding dimension from $2$ to $20$, and graph context each contributes about half the improvement; 2) on WN18RR, {\it OTE} outperforms {\it RotatE} and {\textit{GC-OTE}} achieves the new state-of-the-art results (as far as we know from published papers). 
These results show the effectiveness of the proposed {\it OTE} and graph context for the task of predicting missing links in knowledge graph.

Moreover, {\textit{GC-OTE}} improves more on FB15k-237 than on WN18RR. This is because FB15k-237 has richer graph structure context compared to WN18RR: an average of $19$ edges per node v.s. $2$ edges per node in WN18RR. These results indicate that the proposed method {\it GC-OTE} is more effective on data set with rich context structure information.





\begin{table}[!h]
    \small
    \centering
    \begin{tabular}{c|c|c|c|c}
        \hline
        Model & $d_s$ & MRR & @10 & \#param\\
         \hline
         RotatE-S & - & .330 & .515 & 5.9 \\
         \hline
         RotatE-L & - & .340 & .530 & 29.3 \\
         \hline
         \hline
         OTE & 2 & .327 & .511 & 6.1 \\
         \hline
         OTE & 20 & .355 & .540 & 7.8 \\
         \hline
         \hline
         OTE - scalar & 20 & .352 & .535 & 7.7 \\
         \hline
         LNE & 20 & .354 & .538 & 9.6 \\
         \hline
         \hline
         GC-RotatE-L & - & .354  & .546 & 29.3 \\
         \hline
         GC-OTE & 20 & .367 & .555 & 7.8 \\
         \hline
    \end{tabular}
 \caption{Ablation study on FB15k-237 validation set.} 
    \label{tab:ablation}
\end{table}

\subsubsection{Ablation Study}
Table~\ref{tab:ablation} shows the results of ablation study of the proposed models and compares the number of model parameters with \textit{RotatE} on FB15k-237 validation set. We perform the ablation study with embedding dimension of $400$. The entity embedding dimension for {\it RotatE-S} and {\it RotatE-L} are $400$ and $2000$, respectively.

First we notice that increasing embedding size from $400$ to $2000$ makes {\it RotatE} model size more than quadrupled while the performance gain is very limited (Row 1 and 2 in Table~\ref{tab:ablation}); increasing group embedding size from $2$ to $20$ does not increase the model size of {\it OTE} much, but with nice performance gain (Row $3$ and $4$ in Table~\ref{tab:ablation}). The model size of {\it OTE} is less than one-third of the size of {\it RotatE-L} but with better performance. This shows the effectiveness of the {\it OTE}.

We examine the proposed model in terms of the following aspects:

\noindent\textbf{Impact of sub-embedding dimension}: we fix the embedding dimension as $400$, and increase the sub-embedding dimension $d_s$ from $2$ to $20$, the MRR of \textit{OTE} is improved from 0.327 to 0.355 (See Row 3 and Row 4).  For \textit{RotatE}, the entity is embedded in complex vector space, this is similar to our setting with sub-embedding dimension = 2. Our results show that increasing the sub-dimension with {\it OTE} is beneficial to link prediction. 
    
\noindent\textbf{Impact of orthogonal transform}: we replace the orthogonal transform operation in {\it OTE} with two different settings, 1) removing the diagonal scalar tensor as Eq.~\ref{equ:hr_t_g} (See \textit{OTE-scalar}) and 2) using normal linear transform rather than orthogonal transform (See \textit{LNE}). Both settings lead to MRR degradation. This indicates the proposed orthogonal transform is effective in modeling the relation patterns which are helpful for link prediction.
    
\noindent\textbf{Impact of graph context}: we add the graph context based model to both  {\it OTE} (See \textit{GC-OTE}) and {\it RotatE-L} (See \textit{GC-RotatE-L}). We observe that MRRs are improved for both {\it RotatE-L} and {\it OTE}. This shows the importance of modeling context information for the task of link prediction.

\begin{figure}[t!]
    \includegraphics[width=0.46\textwidth]{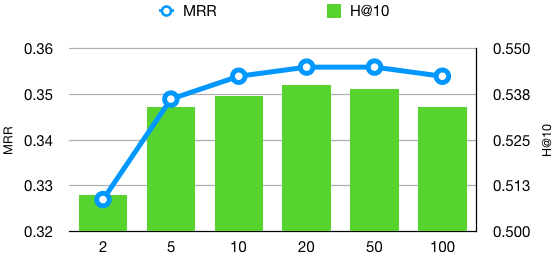}
    \caption{FB15k-237 for {\it OTE} with different sub-embedding dimension.}
    \label{fig:groupdim}
\end{figure}

\noindent\textbf{Sub-embedding dimension size:} in Table \ref{tab:ablation} we show that increasing sub-embedding dimension brings a nice improvement on MRR. Is the larger size always better?
Figure~\ref{fig:groupdim} shows the impact of
$d_s$ on the {\it OTE} performance with the changing of sub-embedding size. We fix the entity embedding dimension as $400$, and vary the sub-embedding size from $2, 5, 10, 20, 50$, all the way to $100$. The blue line and green bar represent MRR and $H@10$ value, respectively. 

From Figure~\ref{fig:groupdim} we observe that, both MRR and Hit@10 are improved and slowly saturated around $d_s$ = 20
The similar experiments are also conducted on WN18RR data set and we find the best sub-embedding dimension is $4$ on WN18RR.

\begin{table}[!h]
    \centering\scriptsize
    \begin{tabular}{c|c|c|c|c|c|c|c}
        \hline
         & &\multicolumn{3}{c|}{\textbf{RotatE-L}} & \multicolumn{3}{c}{\textbf{GC-OTE}} \\
        \cline{3-8}
       Type  & Num. & H &T & A & H &T & A \\
       \hline
       1-to-N & 2255 &.710 & .169 & .440 & .718 &.204 & .461 \\
       \hline
       N-to-1 & 5460 &.156 & .850 & .503 & .209 & .863 & .536 \\
       \hline
       N-to-N & 9763 &.490 & .631 & .561 & .508 & .651 & .579\\
       \hline
    \end{tabular}
    \caption{H@10 from FB15-237 validation set by categories (1-to-N, N-to-1 and N-to-N).}
    \label{tab:categories}
\end{table}

\subsubsection{Error Analysis}
We present error analysis of the proposed model on 1-to-N, N-to-1 and N-to-N relation predictions on FB15k-237. Table~\ref{tab:categories} shows results in terms of Hit@10, where ``Num." is the number of triples in the validation set belonging to the corresponding category, ``H''/``T'' represents the experiment to predict head entity /tail entity, and ``A'' denotes average result for both ``H'' and ``T''.



Assume $c(h,r)$ and $c(r,t)$ are the number of $(h,r)$ and $(r,t)$ pairs appeared in triples from the training set respectively.  A triple $(h,r,t)$ from the validation set is considered as one of the categories in the following:
\begin{equation}
(h,r,t) \!\!= \!\!\begin{cases}
\text{N-to-1},\ \!\!\!\!&\!\!\!\! \text{if } c(h,r)>1 \text{ and } c(r,t)\leq 1 \\
\text{1-to-N},\ \!\!\!\!&\!\!\!\! \text{if } c(h,r)\leq 1 \text{ and } c(r,t)>1 \\
 \text{N-to-N},\ \!\!\!\!&\!\!\!\!  \text{if } c(h,r)>1 \text{ and } c(r,t)>1 \\
\text{other.} & \nonumber\\ 
\end{cases}
\end{equation}

From Table~\ref{tab:categories} we observe that, comparing to \textit{RotatE} large model, the proposed model get better Hit@10 on all cases, especially for the difficult cases when we attempt to predicting the head entity for 1-to-N/N-to-N relation type, and tail entity in N-to-1/N-to-N relation type. The reason is because that in the proposed model, the groupings of sub-embedding relation pairs in {\it OTE} and graph context modeling both are helpful to distinguish N different tails/heads when they share the same (head, rel)/(rel, tail). 



\section{Conclusions}
In this paper we propose a new distance-based knowledge graph embedding for link prediction. It includes two-folds. First, {\it OTE} extends the modeling of {\it RotatE} from 2D complex domain to high dimensional space with orthogonal relation transforms. Second, graph context is proposed to integrate graph structure information into the distance scoring function to measure the plausibility of the triples during training and inference.



The proposed 
approach effectively improves prediction accuracy on the difficult N-to-1, 1-to-N and N-to-N link predictions. 
Experimental results on standard benchmark FB15k-237 and WN18RR show that {\it OTE} improves consistently over {\it RotatE}, the state-of-the-art distance-based embedding model, especially on FB15k-237 with many high in-degree nodes. On WN18RR our model achieves the new state-of-the-art results. This work is partially supported by Beijing Academy of Artificial Intelligence (BAAI).




\clearpage
\appendix

\section{Discussion on the Ability of Pattern Modeling and Inference}
\label{sec:appendix}
It can be proved that {\it OTE} can infer all three types of relation patterns, e.g., symmetry/antisymmetry, inversion and composition patterns.
\subsection{Symmetry/antisymmetry} 
If $e_t=f(r,h)$ and $e_h=f(r,t)$ hold, we have
\begin{eqnarray}
    e_t &=& diag(\exp(s_r))\phi(M_r) \nonumber \\
    &&diag(\exp(s_r))\phi(M_{r})  e_t  \nonumber \\
    \Rightarrow&& \phi(M_r)\phi(M_{r}) = I \nonumber \\
    &&  s_r = 0           \nonumber
\end{eqnarray}{}
In other words, if $\phi(M_r)$ is a symmetry matrix and no scale is applied, the relation is symmetry relation.

If the relation is antisymmetry, e.g., $e_t=f(r,h)$ and $e_h \neq f(r,t)$, we just need to one of the $\phi(M_r(i))$ is not symmetry matrix or $s_r(i) \neq 0$.

\subsection{Inversion} 
If $e_2=f(r_1, e_1)$ and $e_1=f(r_2, e_2)$ hold, we have
\begin{eqnarray}
    e_2 &=& diag(\exp(s_{r_1}))\phi(M_{r_1}) \nonumber \\
    && diag(\exp(s_{r_2}))\phi(M_{r_2})e_2 \nonumber
\end{eqnarray}
In other words, if $diag(\exp(s_{r_1}))\phi(M{r_1})=\phi(M_{r_2})^Tdiag(\exp(-s_{r_2}))$, the relation $r_2$ is inverse relation of $r_1$.

\subsection{Composition} 
If $e_2=f(r_1,e_1)$,  $e_3=f(r_2,e_2)$ and $e_3=f(r_3, e_1)$ hold, we  have
\begin{eqnarray}
   & diag(\exp(s_{r_3})) \phi(M_3) e_1= \nonumber \\
   &   diag(\exp(s_{r_2}))\phi(M_2) \nonumber \\
   &  diag(\exp(s_{r_1}))\phi(M_1) e_1  \nonumber
\end{eqnarray}
It means if $diag(\exp(s_{r_3}))\phi(M_3)$ is equal to $diag(\exp(s_{r_2}))\phi(M_2)diag(\exp(s_{r_1}))\phi(M_1)$ then relation $r_3$ is composition of relation $r_1$ and $r_2$.

\end{document}